\pdfoutput=1

\documentclass[11pt]{article}
\usepackage{multirow} 
\usepackage{acl}

\usepackage{times}
\usepackage{latexsym}
\usepackage{listings}

\usepackage{mdframed} 
\usepackage{listings} 
\usepackage{xcolor} 
\usepackage{pifont}
\newcommand{\cmark}{\textcolor{green}{\checkmark}} 
\newcommand{\xmark}{\textcolor{red}{\ding{55}}}    

\usepackage{booktabs}

\usepackage[T1]{fontenc}

\usepackage[utf8]{inputenc}

\usepackage{microtype}

\usepackage{inconsolata}

\usepackage{amsfonts,amsmath}
\usepackage{graphicx}

\title{Extracting Polymer Nanocomposite Samples from Full-Length Documents}

\author{\textbf{Ghazal Khalighinejad}$^1$, \textbf{Defne Circi}$^2$, \textbf{L.C. Brinson}$^2$, \textbf{Bhuwan Dhingra}$^1$ \\
        $^1$\text{Department of Computer Science, Duke University, USA} \\ $^2$\text{Department of Mechanical Engineering and Materials Science, Duke University, USA} \\
        \{ghazal.khalighinejad, defne.circi, cate.brinson, bhuwan.dhingra\}@duke.edu}

\mdfdefinestyle{PromptStyle}{
  linecolor=black,
  outerlinewidth=1pt,
  roundcorner=5pt,
  innertopmargin=10pt,
  innerbottommargin=10pt,
  innerrightmargin=10pt,
  innerleftmargin=10pt,
  backgroundcolor=gray!10!white
}

\begin{document}
\maketitle
\begin{abstract}
This paper investigates the use of large language models (LLMs) for extracting sample lists of polymer nanocomposites (PNCs) from full-length materials science research papers. The challenge lies in the complex nature of PNC samples, which have numerous attributes scattered throughout the text. 
The complexity of annotating detailed information on PNCs limits the availability of data, making conventional document-level relation extraction techniques impractical due to the challenge in creating comprehensive named entity span annotations.
To address this, we introduce a new benchmark and an evaluation technique for this task and explore different prompting strategies in a zero-shot manner. We also incorporate self-consistency to improve the performance. Our findings show that even advanced LLMs struggle to extract all of the samples from an article. Finally, we analyze the errors encountered in this process, categorizing them into three main challenges, and discuss potential strategies for future research to overcome them.
\end{abstract}

\section{Introduction}
\label{sec:intro}

Research publications are the main source for the discovery of new materials in the field of materials science, providing a vast array of essential data. Creating structured databases from these publications enhances discovery efficiency, as evidenced by AI tools like GNoME~\citep{Merchant2023}. Yet, the unstructured format of journal data complicates its extraction and use for future discoveries~\citep{horawalavithana2022foundation}. Furthermore, the manual extraction of material details is inefficient and prone to errors, underlining the necessity for automated systems to transform this data into a structured format for better retrieval and analysis~\citep{yang2022piekm}.

\begin{figure}
    \centering
    \includegraphics[scale = 0.2]{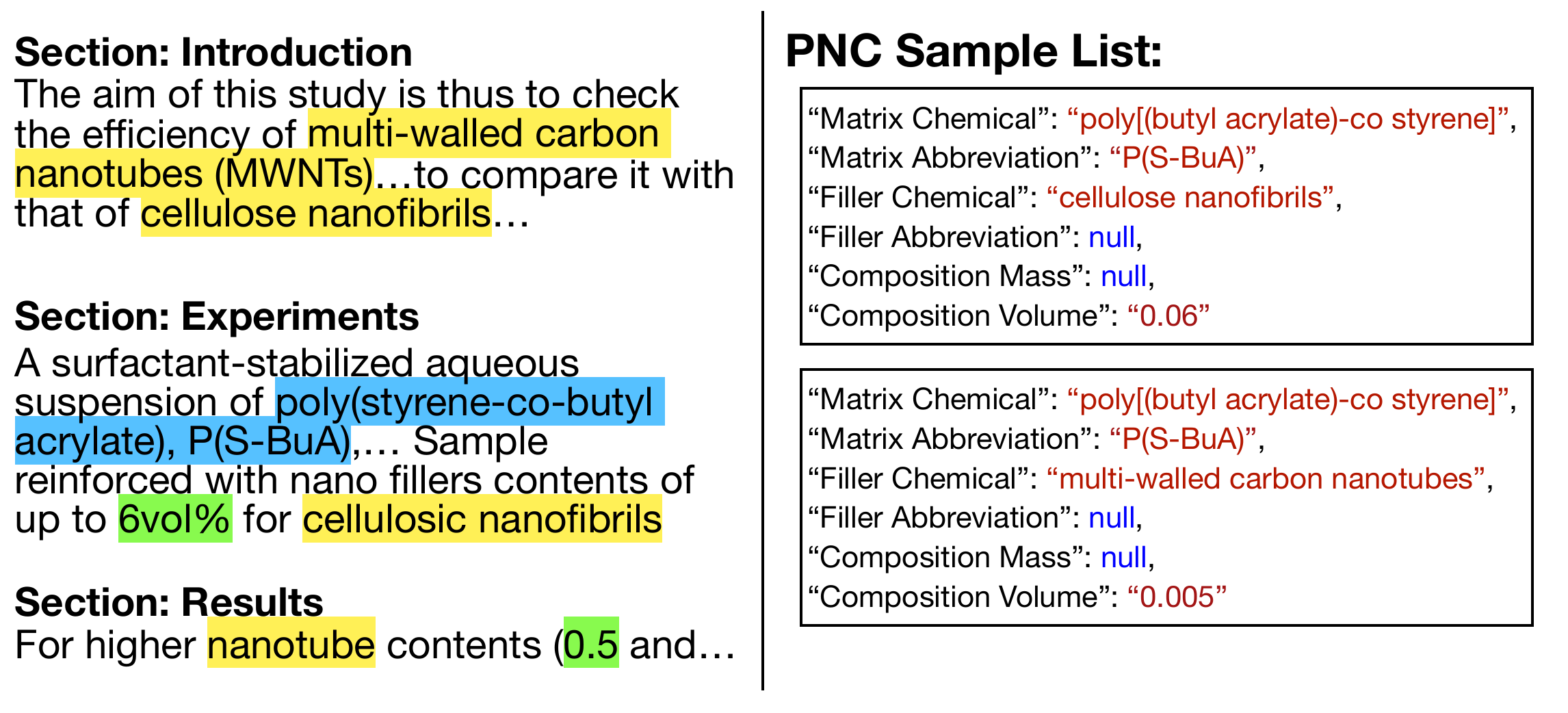}
    \caption{A snippet from a PNC research article~\citep{DALMAS2007829} and the extracted PNC sample list from the NanoMine database.
    Note how information for a single sample is
    extracted from multiple parts of the article text.
    }
    \label{fig:pnc-example}
\end{figure}

Scientific papers on polymer nanocomposites (PNCs) provide essential details on sample compositions, key to understanding their properties. PNCs, which blend polymer matrices with nanoscale fillers, are significant in materials science for their customizable mechanical, thermal, and electrical characteristics. The variety in PNCs comes from different matrix and filler combinations that modify the properties. However, extracting this data poses challenges due to its distribution across texts, figures, and tables, and the complexity of $N$-ary relationships defining each sample. An example in Figure~\ref{fig:pnc-example} illustrates how sample details can be spread over various paper sections.

In this paper, we construct PNCExtract, a benchmark designed for extracting PNC sample lists from scientific articles. PNCExtract focuses on the systematic extraction of $N$-ary relations across different parts of full-length PNC articles, capturing the unique combination of matrix, filler, and composition in each sample. Many works have explored $N$-ary relation extraction from materials science literature~\citep{Dunn2022StructuredIE, song2023matsci, song2023honeybee, xie2023large, cheung2023polyie} and other domains~\citep{giorgi-etal-2022-sequence}. However, these studies primarily target abstracts and short texts, not addressing the challenge of extracting information from the entirety of full-length articles. PNCExtract addresses this by requiring models to process entire articles, identifying information across all sections, a challenge noted by \citet{hira2023reconstructing}. 

Compared to other document-level information extraction (IE) datasets like SciREX~\citep{jain-etal-2020-scirex}, PubMed~\citep{jia-etal-2019-document}, and NLP-TDMS~\citep{hou-etal-2019-identification} which also demand the analysis of entire documents for $N$-ary relation extraction, our dataset marks the first initiative within the materials science domain. This distinction is important due to the unique challenges of IE in materials science, particularly with polymers. The field features a complex nomenclature with chemical compounds and materials having various identifiers such as systematic names, common names, trade names, and abbreviations, all with significant variability and numerous synonyms for single entities~\citep{doi:10.1021/acs.jcim.6b00207}. Furthermore, there is a scarcity of annotated datasets with detailed information, which complicates the creation of effective IE models in this area.

\begin{table}[t!]
\scriptsize
\centering
\begin{tabular}{c|ccccc}
\toprule
\textbf{Task} & \textbf{Doc-level} & \textbf{$N$-ary RE} & \textbf{End-to-End}  \\
\midrule
\multicolumn{4}{c}{\textbf{Materials Domain}} \\
\midrule
PNCExtract & \cmark & \cmark & \cmark \\
PolyIE & \xmark & \cmark & \xmark  \\
\citet{Dunn2022StructuredIE} & \xmark & \cmark & \cmark  \\
\citet{xie2023large} & \cmark & \xmark & \cmark \\
MatSci-NLP & \xmark & \xmark & \xmark \\
\midrule
\multicolumn{4}{c}{\textbf{Other Domains}} \\
\midrule
PubMed & \cmark & \cmark & \xmark  \\
SciREX & \cmark & \cmark & \xmark  \\
NLP-TDMS & \cmark & \cmark & \xmark \\
\bottomrule
\end{tabular}
\caption{Comparison of PNCExtract with other Information Extraction (IE) approaches within the materials science domain~\citep{song2023matsci, cheung2023polyie, Dunn2022StructuredIE, xie2023large} and across various other scientific domains~\citep{jain-etal-2020-scirex,jia-etal-2019-document, hou-etal-2019-identification}. ``End-to-End'' indicates that, unlike previous methods that require task-specific supervision (e.g., named entity recognition, coreference resolution), PNCExtract relies on end-to-end supervision only.
}
\label{tab:comparison-dataset}
\end{table}

In light of these challenges, our dataset is designed for a generative task to navigate the complexities of fully annotating entire PNC papers, which involve annotating named entity spans, coreferences, and negative examples (entity pairs without a relation). The complexity of PNC papers, due to their various entities and samples, makes manual annotation both time-consuming and prone to errors. Consequently, encoder-only models, which require extensive annotations, fall short for our purposes. In Table~\ref{tab:comparison-dataset}, we compare PNCExtract with previous IE approaches in the scientific domain.

We introduce a dual-metric evaluation system comprising a partial metric for detailed analysis of each attribute within an $N$-ary extraction and a strict metric for assessing overall accuracy. This approach distinguishes itself from prior works in materials science, which either focused on the evaluation of binary relations~\citep{Dunn2022StructuredIE, xie2023large, song2023matsci, wadhwa-etal-2023-revisiting} or used strict evaluation criteria~\citep{cheung2023polyie} without recognizing partial matches.

We further explore different prompting strategies, including one that aligns with the principles of Named Entity Recognition (NER) and Relation Extraction (RE) which involves a two-stage pipeline, as well as an end-to-end method to directly generate the $N$-ary object. We find that the E2E approach works better in terms of both accuracy and efficiency.
Moreover, we present a simple extension to the self-consistency technique~\citep{wang2023selfconsistency} for list-based predictions. Our findings demonstrate that this approach improves the accuracy of sample extraction. Since the extended length of articles often exceeds the context limits of some LLMs, we also explore condensing them through a dense retriever~\citep{ni-etal-2022-large} to extract segments most relevant to specific queries. Our findings indicate that condensing documents generally enhances accuracy. Since existing document-level IE models~\citep{jain-etal-2020-scirex, zhong-chen-2021-frustratingly} are not suited for our task, we employ GPT-4 with our E2E prompting on the SciREX dataset and benchmark it against the baseline model. Our analysis shows that GPT-4, even in a zero-shot setting, outperforms the baseline models that were trained with extensive supervision.

Lastly, we discuss three primary challenges encountered when using LLMs for PNC sample extraction. Code for reproducing all experiments is available at \url{https://github.com/ghazalkhalighinejad/PNCExtract}.

\section{PNCExtract Benchmark}

In this section, we first describe our dataset, including the problem definition, and the dataset preparation. Then we describe our evaluation method for the described task.

\subsection{Problem Definition}

\label{sec:task}
We define our dataset as \( \mathcal{D} = \{D_1, D_2, \ldots, D_{193}\} \), where each \( D_i \) is a peer-reviewed paper included in our study. Corresponding to each paper \( D_i \), there is an associated list of samples \( \mathcal{S}_i \), comprising various PNC samples. Formally, \( \mathcal{S}_i \) is defined as \( \mathcal{S}_i = \{s_{i1}, s_{i2}, \ldots, s_{in_i}\} \), where \( s_{ij} \) represents the \( j \)-th PNC sample in the sample list of the \( i \)-th paper, and \( n_i \) denotes the total number of PNC samples in \( \mathcal{S}_i \). Each sample \( s_{ij} \) is a JSON object with six entries: Matrix Chemical Name, Matrix Chemical Abbreviation, Filler Chemical Name, Filler Chemical Abbreviation, Filler Composition Mass, and Filler Composition Volume. Table~\ref{tab:nonnull_samples} presents the count of samples with each attribute marked as non-null. The primary task involves extracting a set of samples \( \hat{\mathcal{S}}_i \) from a given paper \( D_i \).
\begin{table}[!htbp]
\centering
\small
\begin{tabular}{lcc}
\toprule
Attribute & Number of Samples \\ \midrule
Matrix Chemical Name & 1052 \\
Matrix Chemical Abbreviation & 864 \\
Filler Chemical Name & 1052 \\
Filler Chemical Abbreviation & 819 \\
Filler Mass & 624 \\ 
Filler Volume & 407 \\ 
\bottomrule
\end{tabular}
\caption{Number of total samples for which each of the attributes is non-null.}
\label{tab:nonnull_samples}
\end{table}

\begin{table}[!htbp]
\centering
\small
\begin{tabular}{lccc}
\toprule
\textbf{Statistics} & Paper Length & \#Samples/Doc \\
\midrule
Avg. & 6965 & 6 \\
Med. & 6734 & 4 \\
Min. & 238  & 1 \\
Max. & 16355 & 50\\
\bottomrule
\end{tabular}
\caption{Statistical summary of paper lengths and number of samples per document. Paper length is measured in tokens.}
\label{tab:paper_lengths}
\end{table}

\subsection{Dataset Preparation}

\paragraph{NanoMine Data Repository}
We curate our dataset using the NanoMine data repository~\citep{10.1063/1.5046839}. NanoMine is a PNC data repository structured around an XML-based schema designed for the representation and distribution of nanocomposite materials data. The NanoMine database is manually curated using Excel templates provided to materials researchers. NanoMine database currently contains a list of $240$ full-length scholarly articles and their corresponding PNC sample lists. While NanoMine includes various subfields, our study focuses on the ``Materials Composition'' section. This section comprehensively details the characteristics of constituent materials in nanocomposites, including the polymer matrix, filler particles, and their compositions (expressed in volume or weight fractions). The reason for this focus is that determining which sample compositions are studied in a given paper is the essential first step toward identifying and understanding more complex properties of PNCs. Out of the $240$ articles, we focus on $193$ and disregard the rest due to having inconsistent format (see Appendix~\ref{app:process-nanomine}). These $193$ articles contain a total of $1052$ samples. For each sample, we retain $6$ out of the $43$ total attributes in the Materials Composition of NanoMine (see Appendix~\ref{app:curation} for details). 

Document-level information extraction requires understanding the entire document to accurately annotate entities, their relations, and saliency. These make the annotation of scientific articles time-consuming and prone to errors. We found that NanoMine also contains errors. Given the challenge of reviewing all $1052$ samples and reading through $193$ articles, we adopted a semi-automatic approach to correct samples. Specifically, for an article, we consider both the predicted and ground truth sample list of a document. Using our partial metric (detailed in Section~\ref{sec:partial}), we match predicted samples with their ground-truth counterparts and assign a similarity score to each pair. Matches are classified as exact, partial, or unmatched—either true samples or predictions. We then focus on re-annotating samples with the most significant differences between prediction and ground truth, especially those partial matches with lower similarity scores and unmatched samples. This method accelerates re-annotation by directing annotators towards specific attributes and samples based on GPT-4 predictions. Following this strategy, we made three types of adjustments to the dataset: deleting $20$ samples, adding $15$, and editing $19$ entities.\footnote{This work includes contributions from polymer experts, under whose mentorship all authors received their training.}(See Appendix~\ref{app:re-annot} for details).

\subsection{Evaluation Metrics}
Our task involves evaluating the performance of our model in predicting PNC sample lists. One natural approach, also utilized by \citet{cheung2023polyie}, is to verify if there is an exact match between the predicted and the ground-truth samples. This method, however, has a notable limitation, particularly due to the numerous attributes that define a PNC sample. Under such strict evaluation criteria, a predicted sample is considered entirely incorrect if even one attribute is predicted inaccurately, which can be too strict considering the complexity and attribute-rich nature of PNC samples.

Hence, we also propose a partial metric which rewards predicted samples
for partial matches to a ground truth sample.
However, computing such a metric first requires identifying the optimal
matching between the predicted and ground truth sample lists,
for which we employ a maximum weight bipartite matching algorithm.
This approach acknowledges the accuracy of a prediction even if not all attributes are perfectly matched.

Additionally, we also apply a strict metric, similar to the approach of \citet{cheung2023polyie}, where a prediction is considered correct only if it perfectly matches with the ground truth across all attributes of a PNC sample.

\paragraph{Standardization of Prediction}
To accurately calculate the partial and strict metrics, standardizing predictions is essential. The variability in polymer name expressions in scientific literature makes uniform evaluation challenging. For example, ``silica'' and ``silicon dioxide'' are different terms for the same filler. Our dataset uses a standardized format for chemical names. To align the predicted names with this standard, we use resources by \citet{Hu2021}, which list $89$ matrix names with their standard names, abbreviations, synonyms, and trade names, as well as $159$ filler names with their standard names. We standardize predicted chemical names by matching them to the closest names in these lists and converting them to their standard forms. 
Furthermore, our dataset exclusively uses numerical values to represent compositions (e.g., a composition of ``$0.5 \text{vol.}\%$'' should be listed as ``$0.005$''). Predictions in percentage format (like ``$0.5 \text{vol.}\%$'') are thus converted to the numerical format to align with the dataset's representation.

\paragraph{Attribute Aggregation}

Our evaluation incorporates an attribute aggregation method. For both the ``Matrix'' and ``Filler'' categories, a prediction is considered accurate if the model successfully identifies either the chemical name or the abbreviation. For the ``Composition'', a correct prediction may be based on either the ``Filler Composition Mass'' or the ``Filler Composition Volume''. This approach allows for a broader assessment, capturing any correct form of attribute identification without focusing on the finer details of each attribute.

\paragraph{Partial-F1} 
\label{sec:partial}
This metric employs the F$_1$ score in its calculation, which proceeds in two steps. Initially, an accuracy score is computed for each pair of predicted and ground truth samples where we compute the fraction of matches in the <Matrix, Filler, Composition> trio across the two samples. 
This process results in $\hat{k} \times k^{}$ score combinations, where $\hat{k}$ and $k^{}$ represent the counts of predicted and ground truth samples. The next step involves translating these comparisons into an assignment problem within a bipartite graph. Here, one set of vertices symbolizes the ground truth samples, and the other represents the predicted samples, with edges denoting the F$_1$ scores between pairs. The objective is to identify a matching that optimizes the total F$_1$ score,
which can be computed using
the Kuhn-Munkres algorithm~\citep{https://doi.org/10.1002/nav.3800020109}.
in $O(n^3)$ time (where $n = max(\hat{k}, k^{})$).
Note that if $\hat{k} \neq k^{}$, a one-to-one match for each prediction may not be necessary. Once matching is done, we count all the correct, false positive, and false negative predicted attributes (the attributes of all the unmatched predicted samples and ground-truth samples are considered false positives and false negatives, respectively). Subsequently, we calculate the micro-average Precision, Recall, and F$_1$. 


\paragraph{Strict-F1} 
For a stricter assessment, a sample is labeled correct only if it precisely matches one in the ground truth. Predictions not in the ground truth are false positives, and missing ground truth samples are false negatives. This metric emphasizes exact match accuracy.

\section{Modeling Sample List Extractions from Articles with LLMs}
\label{sec:model}
As mentioned in Section~\ref{sec:intro}, our dataset is designed for a generative task, making encoder-only models unsuitable for two main reasons. First, these models demand extensive annotations, such as named entity spans, coreferences, and negative examples, a process that is both time-consuming and error-prone. Second, encoder-only models struggle with processing long documents efficiently. While some studies have successfully used these models for long documents~\citep{jain-etal-2020-scirex}, they had access to significantly larger datasets. Our dataset, however, contains detailed domain-specific information, making it challenging to obtain a similarly extensive dataset.

\begin{figure*}
    \centering
    \includegraphics[scale = 0.22]{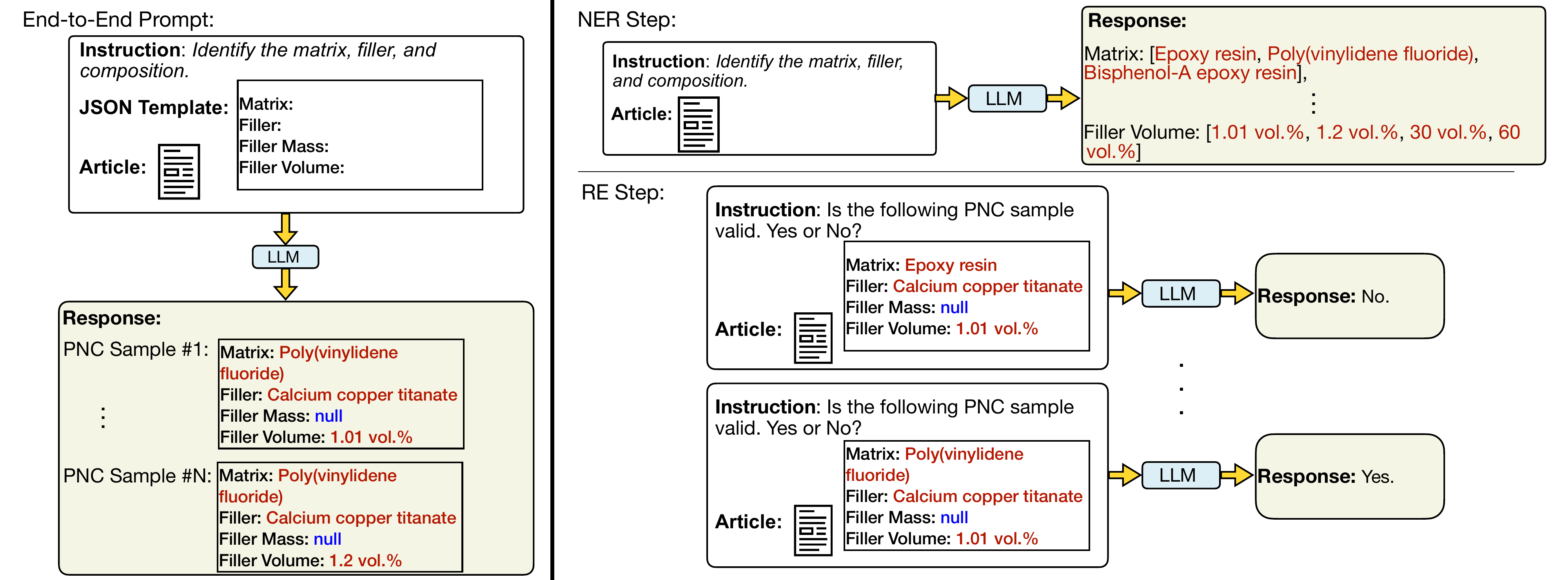}
    \caption{Two prompting strategies for PNC sample extraction with LLM are presented. On the left, the end-to-end (E2E) approach uses a single prompt to directly extract PNC samples. On the right, the NER+RE approach first identifies relevant entities and then classifies their relations through yes/no prompts to validate PNC samples.}
    \label{fig:nerre-prompt}
\end{figure*}

Consequently, within a zero-shot context\footnote{The context length is prohibitive for attempting few-shot approaches.}, we explore two prompting methods: Named Entity Recognition plus Relation Extraction (NER+RE) and an End-to-End (E2E) approach.

\subsection{NER+RE Prompt}
\label{sec:ner+re prompt}
Building on previous research \citep{peng-etal-2017-cross, jia-etal-2019-cross, viswanathan-etal-2021-citationie}, which treated $N$-ary relation extraction as a binary classification task, our NER+RE method treats RE as a question-answering process, following the approach in \citet{zhang-etal-2023-aligning}. This process is executed in two stages. Initially, the model identifies named entities within the text. Subsequently, it classifies $N$-ary relations by transforming the task into a series of yes/no questions about these entities and their relations. For evaluation, we apply only the strict metric, as the partial metric is not suitable in this binary classification context.\footnote{While partial evaluation is theoretically possible by considering all potential samples identified in the NER step, such an approach would yield limited insights.}

The NER+RE approach becomes computationally expensive during inference, especially as the number of entities increases. This leads to an exponential growth in potential combinations, expanding the candidate space for valid compositions and consequently extending the inference time.

\subsection{End-to-End Prompt}
\label{sec:e2e prompt}

To address this challenge, we develop an End-to-End (E2E) prompting strategy that directly extracts JSON-formatted sample data from articles. This method is designed to efficiently handle the complexity and scale of extracting $N$-ary relations from scientific texts, bypassing the limitations of binary classification frameworks in this context. 

\subsection{Self-Consistency}
The self-consistency method~\citep{wang2023selfconsistency}, aims to enhance the reasoning abilities of LLMs. Originally, this method relied on taking a majority vote from several model outputs. For our purposes, since the output is a set of answers rather than a single one, we apply the majority vote principle to the elements within these sets.

We generate $t$ predictions from the model, each at a controlled temperature of $0.7$. Our objective is to identify which samples appear frequently across these multiple predictions as a sign of higher confidence from the model.

During evaluation, each model run generates a list of predicted samples from a specific paper. We refer to each list as the $k$-th prediction, denoted $S_k = \{a^k_1, a^k_2, ..., a^k_m\}$. For each predicted element $a^i_j$, we determine its match score $\text{match}^i_j$, by counting how frequently it appears across all predictions $\{S_1, S_2,...,S_t\}$. This score can vary from $1$, meaning it appeared in only one prediction, to $t$, indicating it was present in all predictions.

We then apply a threshold $\alpha$ to filter the samples. Those with a $\text{match}^i_j$ at or above $\alpha$ are retained, as they were consistently predicted by the model. Samples falling below this threshold suggest less confidence in the prediction and are removed.

\subsection{Condensing Articles with Dense Retrieval}
LLMs, such as LLaMA2 with its token limit of $4,096$, face challenges in maintaining performance with longer input lengths. Recent advancements have extended these limits \citep{longchat2023, tworkowski2023focused}; however, an increase in input length often leads to a decline in model performance. This raises the question of whether condensing articles could serve as an effective strategy to address such limitations. We, therefore, employ the Generalizable T5-based Dense Retrievers (GTR-large) \citep{ni-etal-2022-large} to retrieve relevant parts of the documents.

 This process involves dividing each document $C_i$ into segments ($\{C_{i1}, C_{i2},...,C_{iN}\}$) and formulating four queries ($Q_j$) to extract targeted information regarding an entity.\footnote{The queries are: ``What chemical is used in the polymer matrix?'', ``What chemical is used in the polymer filler?'', ``What is the filler mass composition?'', and ``What is the filler volume composition?''.} On average, each segment consists of 60 tokens.
 We then calculate the similarity between each pair of segments and queries ($C_{ik}$, $Q_j$). For every query $Q_j$, we select the top $k$ segments ($TopK(Q_j, C_i)$) based on their similarity scores. These top segments from all four queries are then combined to form a condensed version of the original document (\(\bigcup_{j=1}^{4} TopK(Q_j, C_i)\)).

\section{Experiments}

\subsection{Benchmarking LLMs on PNCExtract}
\paragraph{Models} In our experiments, we employ LLaMA-7b-Chat \citep{touvron2023llama}, LongChat-7B-16K \citep{longchat2023}, Vicuna-7B-v1.5 and Vicuna-7B-v1.5-16K \citep{vicuna2023}, and GPT-4 Turbo \citep{openai2023gpt4}. The LongChat-7B-16K and Vicuna-7B-16K models are fine-tuned for context lengths of 16K tokens, and GPT-4 Turbo for 128K tokens.

\paragraph{Setup} We divide our dataset into $52$ validation articles and $141$ test articles. We assess the performance using micro average Precision, Recall, and F1 scores, considering both strict and partial metrics at the sample and property levels. We also compare two different prompting strategies NER+RE and E2E. Moreover, we consider the self-consistency technique.

\subsubsection{Results}
In Table~\ref{tab:main} we report the partial and strict metrics for multiple models and settings. We report the best results for each model in the condensed paper setting, selected across different $k = \{5, 10, 30\}$, which correspond to average token counts per document of $790$, $1420$, and $3310$, respectively. Further details on the results across various levels of document condensation are available in the Appendix~\ref{app:all_results}. The results highlight several key observations:

\begin{table}[ht]
\centering
\small

\begin{tabular}{@{}lrrrrrrrr@{}}
\toprule
\textbf{Model} & \multicolumn{3}{c}{\textbf{Strict}} & \multicolumn{3}{c}{\textbf{Partial}} \\ 
      & \textbf{P} & \textbf{R}& \textbf{F$_1$}& \textbf{P}& \textbf{R}& \textbf{F$_1$}  \\
\midrule
\multicolumn{7}{c}{Condensed Papers} \\
\midrule
LLaMA2 C      & 21.7 & 0.6 & 1.2   & 60.0 & 1.5 & 3.0    \\
Vicuna     & 5.8 & 2.6 & 3.6   & 49.9 & 19.5 & 28.1  \\ 
Vicuna-16k      & 17.7 & 5.9 & 8.9   & 60.4 & 19.9 & 29.9  \\
LongChat     & 6.6 & 3.5 & 4.6   & 47.3 & 24.4 & 32.2        \\
GPT-4     & 43.6 & 32.0 & 36.9  & 64.5 & \textbf{47.7} & 54.8  \\ 
\midrule
\multicolumn{7}{c}{Full Papers} \\
\midrule
Vicuna-16k   & 18.4 & 1.5 & 2.7   & 65.7 & 4.6 & 8.5                   \\
LongChat   & 5.4 & 4.2 & 4.7   & 36.6 & 29.6 & 32.7  \\
GPT-4       & 44.8 & 30.2 & 36.0   & 64.9 & 43.8 & 52.3  \\
GPT-4 (NR)         & 28.4 & \textbf{37.2} & 32.2   & - & - & -   \\   
GPT-4 + SC    & \textbf{51.6} & 31.1 & \textbf{38.8}   & \textbf{73.5} & 43.8 & \textbf{54.9} 
\\
\bottomrule
\end{tabular}
\caption{Precision, Recall, and F$_1$ of different LLMs on condensed and full papers using strict and partial metrics. The table includes GPT-4 Turbo with different prompting methods (NER+RE, E2E, and E2E with self-consistency [SC]). ``NR'' denotes NER+RE prompting. ``LLaMA2 C'' represents the LLaMA2-7b-chat model. Models with limited context lengths are evaluated only in the condensed paper scenario.
}
\label{tab:main}
\end{table}

\begin{figure}[ht]
    \centering
    \includegraphics[scale = 0.2]{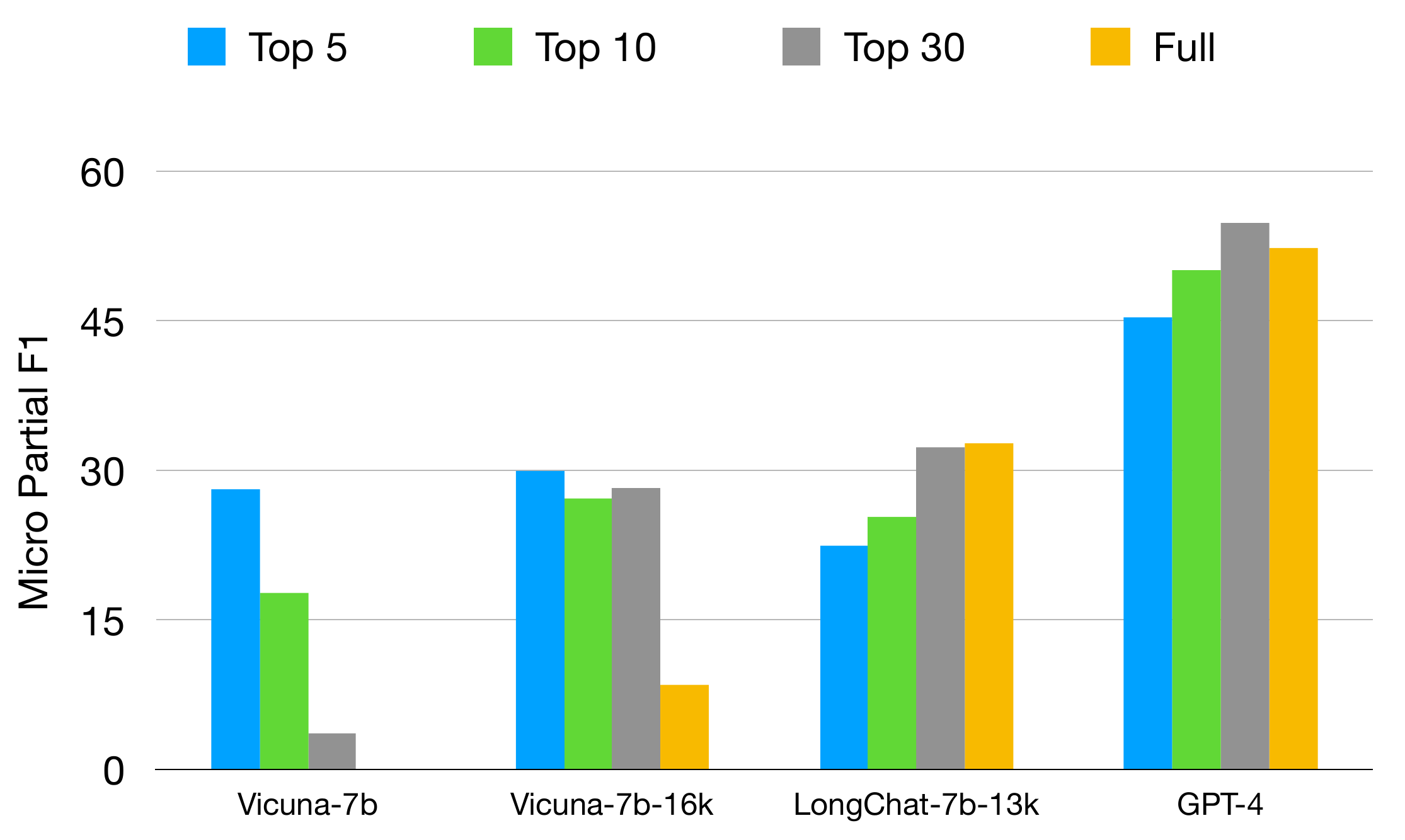}
    \caption{Comparison of Micro Partial F1 Scores Across Different Models and Document Lengths. ``Top 5'', ``Top 10'', and ``Top 30'' indicate document summaries retrieved with $k$ set to $5$, $10$, and $30$ respectively.}
    \label{fig:doclength}
\end{figure}

\paragraph{Effect of Document Length on the Performance}
 Table~\ref{tab:main} demonstrates that shortening documents proves beneficial in most cases. Additionally, Figure~\ref{fig:doclength} shows the trend of partial F$_1$ scores as document length increases. We observe that GPT-4's performance decreases in extremely shortened settings but is optimal when documents are shortened to the top $30$ segments. This indicates that while reducing document length is beneficial, excessive shortening may result in the loss of sample information.
Additionally, Table~\ref{tab:bootstrap} provides bootstrap analysis from $1000$ resamplings, indicating that GPT-4 Turbo has a higher mean F$_1$ score on shorter full-length documents.
\begin{table}[ht]
\centering
\small
\begin{tabular}{@{}lccc@{}}
\toprule
Length Interval & Mean F$_1$ & SD & 95\% CI \\ \midrule
(0, 8000) & 44.2 & 04.0 & (35.2, 51.2) \\
(8000, 20000) & 35.2 & 05.7 & (24.4, 46.7) \\ \bottomrule
\end{tabular}
\caption{Comparison of mean F$_1$ scores, standard deviations, and 95\% confidence intervals for different token length intervals.}
\label{tab:bootstrap}
\end{table}

\paragraph{E2E vs. NER+RE:}
The E2E prompting method shows better performance compared to the NER+RE approach, which is attributed to the higher precision of E2E. Furthermore, the inference time of the GPT-4 Turbo (E2E) is $28$ sec/article, faster than $45$ sec/article for GPT-4 Turbo (NER+RE).

\paragraph{Impact of Self-consistency on PNC Sample Extraction:}
To optimize the application of self-consistency, we first determine the most effective number of predictions to sample and the optimal value for $\alpha$ on the validation set (see Appendix~\ref{app:sc}). Based on that, we employ $\alpha = 3$ and $8$ predictions on the test set. Table~\ref{tab:main} shows that self-consistency enhances the strict and partial F$_1$.

\paragraph{Influence of the Partial Metric}
Adopting the partial metric has several advantages. First, it helps identify specific challenge areas. For example, in Table~\ref{tab:attributes}, we show the model faces the most challenges in accurately predicting Composition. Furthermore, human annotations for PNC samples are often error-prone~\citep{Himanen_2019, 10.1007/978-3-030-62466-8_10}, hence one potential use of an LLM like GPT-4 would be to identify errors and send them back for re-annotation. The partial metric can help prioritize which samples to re-annotate. 

\begin{table}[ht]
\small
\centering
\begin{tabular}{lrrr}
\toprule
Attributes & P & R & F$_1$ \\
\midrule
Matrix & 50.2 & 23.5 & 32.1  \\
Filler & 53.1 & 25.0 & 34.0  \\
Composition & 44.4 & 20.4 & 28.0  \\
\bottomrule
\end{tabular}
\caption{Micro average precision, recall, and F$_1$ across the attributes.}
\label{tab:attributes}
\end{table}

\subsection{Comparing with Baselines}
Previous works on document-level $N$-ary IE~\citep{jain-etal-2020-scirex, jia-etal-2019-document, hou-etal-2019-identification}, have relied on encoder-only models, making them unsuitable for our specific task. For comparative purposes, we prompt GPT-4 on the SciREX dataset \citep{jain-etal-2020-scirex}, which comprises $438$ annotated full-length machine learning papers. As shown in Table~\ref{tab:scirex}, when prompted in a zero-shot, end-to-end manner, GPT-4 Turbo outperforms the baseline methods. Note that the baseline model, trained on $300$ papers, received extensive supervision in the form of mention, coreference, binary relation, and salient mention identification.
This suggests that we would need to expend a large amount of annotation effort on PNCExtract to build a supervised pipeline comparable to
the zero-shot GPT-4 approach presented here.
\begin{table}[ht]
\small
\centering
\begin{tabular}{lrrr}
\toprule
Model & Prec. & Rec. & F$_1$ \\
\midrule
\citet{jain-etal-2020-scirex} & 0.7 & 17.3 & 0.8  \\
Zero-shot GPT-4 & 5.0 & 8.5 & \textbf{5.5}  \\
\bottomrule
\end{tabular}
\caption{Micro average precision, recall, and F$_1$. The baseline results are taken from the referenced paper.}
\label{tab:scirex}
\end{table}

\begin{figure*}
    \centering
    \includegraphics[scale = 0.25]{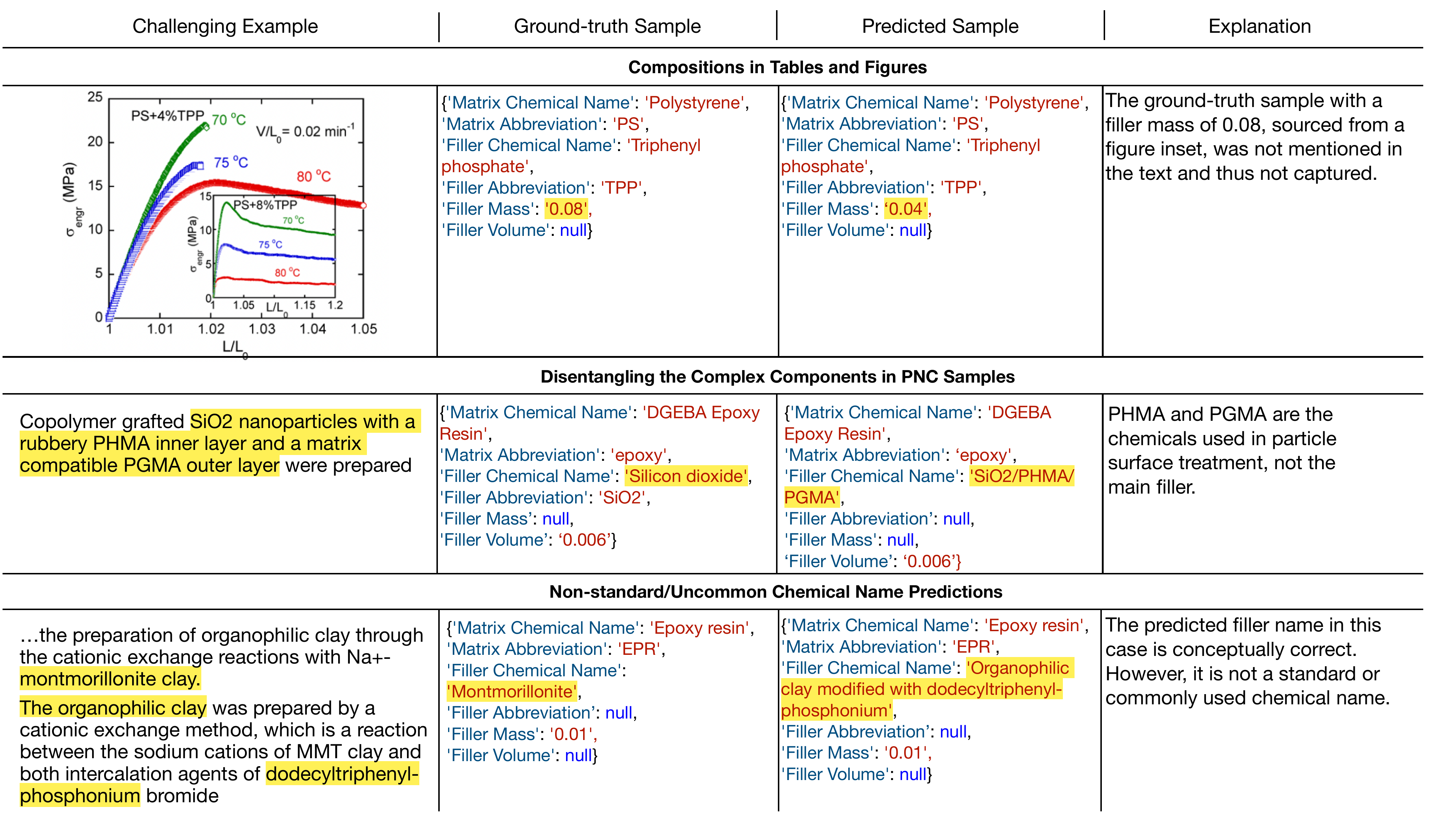}
    \caption{Examples of challenges for LLMs, showcasing three categories of challenges encountered in capturing accurate PNC sample compositions. Each row demonstrates a specific challenge, the ground-truth sample, the model's prediction, and a brief explanation of the issue."}
    \label{fig:examples_casestudy}
\end{figure*}

\subsection{Analysis of Errors}
\label{sec:analysis}
Accurately extracting PNC samples is a complex task, and even state-of-the-art LLMs fail to capture all the samples. We find that out of $1052$ ground-truth samples, $773$ were not identified in the model's predictions. Furthermore, $364$ of the $664$ predictions were incorrect. This section discusses three categories of challenges faced by current models in sample extraction and proposes potential directions for future improvements.

\paragraph{Compositions in Tables and Figures}
NanoMine aggregates samples from the literature, including those presented in tables and visual elements within research articles. As demonstrated in the first example of Figure~\ref{fig:examples_casestudy}, a sample is derived from the inset of a graph. Our present approach relies solely on language models. Future research could focus on advancing models to extract information from both textual and visual data.

\paragraph{Disentangling the Complex Components in PNC Samples}
The composition of PNC includes a variety of components such as hardeners and surface treatment agents. A common issue in our model’s predictions is incorrectly identifying these auxiliary components as the main attributes. For example, the second row in Figure~\ref{fig:examples_casestudy} shows the model predicting the filler material along with its surface treatments instead of recognizing the filler by itself. 

\paragraph{Non-standard/Uncommon Chemical Name Predictions}
The expression of chemical names is inherently complex, with multiple names often existing for the same material. In some cases, predicted chemical names are conceptually accurate yet challenging to standardize. This suggests the necessity for more sophisticated approaches that can handle the diverse and complex representations of chemical compounds. The third example in Figure~\ref{fig:examples_casestudy} shows an example of this.

\section{Related Work}
Early works have focused on training models specifically for the tasks of NER and RE. Building on this, recently \citet{wadhwa-etal-2023-revisiting} and \citet{wang2023gptner} show that LLMs can effectively carry out these tasks through prompting. 

In the specific area of models trained on a materials science corpus, MatSciBERT~\citep{Gupta2022} employs a BERT~\citep{devlin2018bert} model trained specifically on a materials science corpus. \citet{song2023honeybee} further developed HoneyBee, a fine-tuned Llama-based model for materials science. MatSciBERT was not applicable to our task, as detailed in Section~\ref{sec:model}, and HoneyBee's model weights were not accessible during our research phase. Other contributions in this field include studies by \citet{Shetty_2023}, \citet{doi:10.1080/27660400.2021.1899456}, and \citet{osti_1558659}, focusing on the extraction of polymer-related data from scientific articles.

Similar to \citet{Dunn2022StructuredIE}, \citet{xie2023large}, \citet{tang2023does} and \citet{cheung2023polyie} our study also focuses on extracting $N$-ary relations from materials science papers. However, our approach diverges in two significant aspects: we analyze full-length papers, not just selected sentences, and we extend our evaluation to partial assessment of $N$-ary relations, rather than limiting it to binary assessments.

\section{Discussion and Future Works}
We introduced PNCExtract, a benchmark focused on the extraction of PNC samples from full-length materials science articles. To the best of our knowledge, this is the first benchmark enabling detailed $N$-ary IE from full-length materials science articles. We hope that this effort encourages further research into generative end-to-end methods for scientific information extraction from full-length documents. Future investigations should also consider more advanced techniques for condensing entire scientific papers. To overcome the challenges in PNC sample extraction discussed in Section~\ref{sec:analysis}, future studies could investigate multimodal strategies that integrate text and visual data. Additionally, experimenting fine-tuning methods could lead to more precise chemical name generation.

\section{Limitation}
Although our dataset comprises samples derived from figures within the papers, the current paper is confined to the assessment of language models exclusively. We acknowledge that incorporating multimodal models, which can process both text and visual information, has the potential to enhance the results reported in this paper. 

Furthermore, despite our efforts to correct NanoMine, another limitation of our study is the potential presence of inaccuracies within the dataset.

Additionally, our paper selectively examines a subset of attributes from PNC samples. Consequently, we do not account for every possible variable, such as ``Filler Particle Surface Treatment.'' This limited attribute selection means we do not distinguish between otherwise identical samples when this additional attribute could lead to differentiation. Acknowledging this, including a broader range of attributes in future work could lead to the identification of a more diverse array of samples.

\section{Ethics Statement}
We do not believe there are significant ethical issues associated with this research.

\appendix
\section{Processing NanoMine}
\label{app:process-nanomine}
\begin{figure}[ht]
    \centering
    \includegraphics[scale = 0.3]{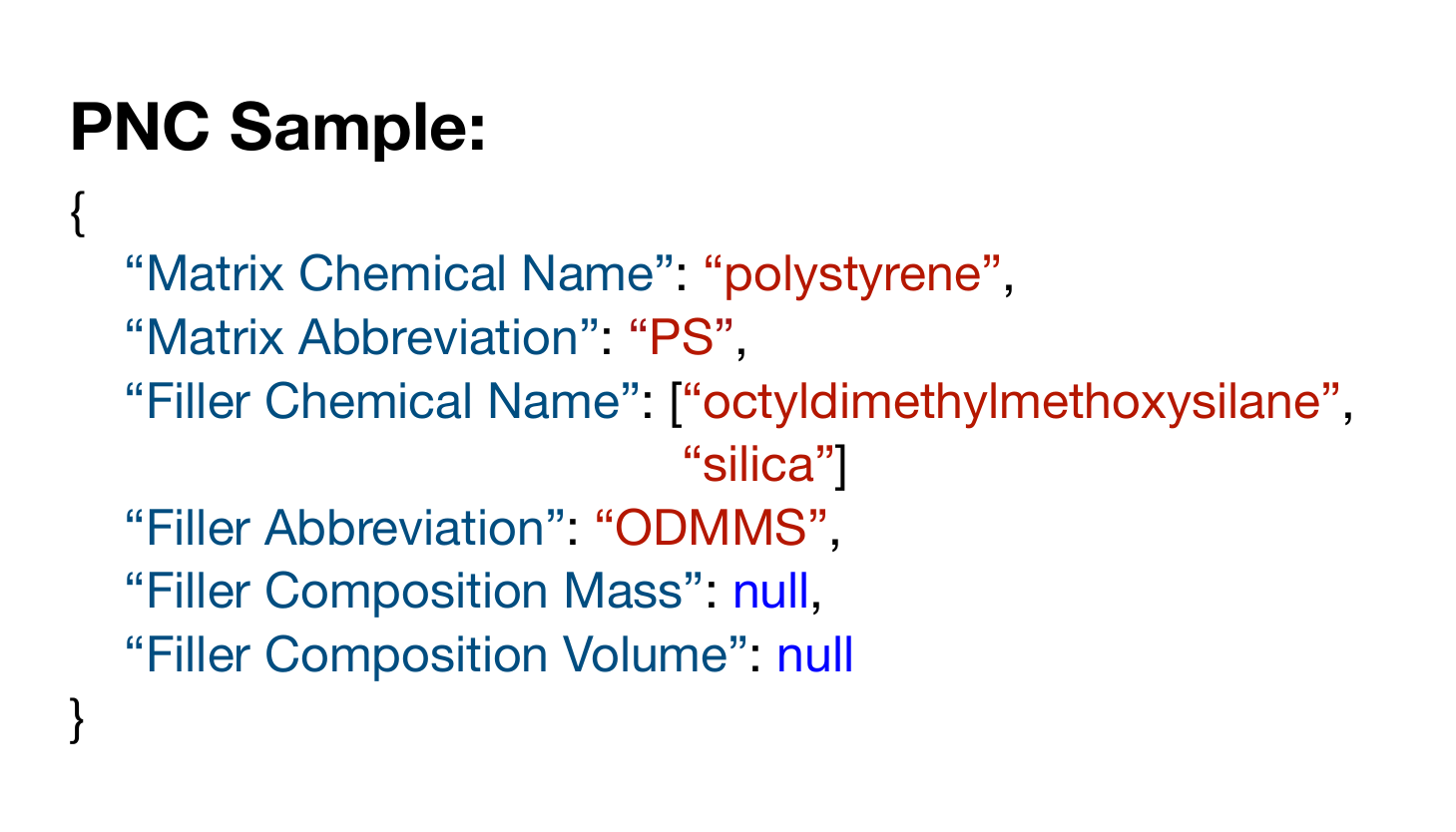}
    \caption{An inconsistent sample in NanoMine that we exclude from our dataset.}
    \label{fig:removed-sample}
\end{figure}

In the sample composition section of NanoMine, various attributes describe the components of a sample. For our analysis, we focus on six specific attributes. Nonetheless, we encounter instances where the formatting in NanoMine is inconsistent. We excluded those articles. This is because our data processing and evaluation require a uniform structure. For example, in Figure~\ref{fig:removed-sample}, we identify an example of an inconsistency where the ``Filler Chemical Name'' is presented as a list rather than a single value, which deviates from the standard JSON format we expect. This inconsistency makes the sample incompatible with our dataset's format, leading to its removal from our analysis.

\section{Dataset Curation and Cleaning}
\label{app:curation}
During our curation process, we selectively disregard certain attributes from NanoMine based on three criteria:

\begin{itemize}
    \item Complexity in Extraction and Evaluation: Attributes that cannot be directly extracted with a language model or evaluated are disregarded. For example, intricate descriptions (such as ``an average particle diameter of 10 um'') are excluded due to their complexity in evaluation.
    \item Rarity in the Dataset: We also disregard attributes infrequently occurring in NanoMine. For instance, ``Tacticity'' is noted in only $0.05\%$ of samples. This rarity might stem from either its infrequent mention in research papers or oversights by annotators.
    \item Relative Importance: Attributes that are less important for our analysis, such as ``Manufacturer Or Source Name'', are also excluded. Our focus is on extracting attributes that are most relevant for identifying a nanocomposite sample.
    
\end{itemize}

\section{Terms of Use}
We used OpenAI (gpt-4 and gpt-4-1106-preview), LLaMA2, LongChat, and Vicuna models, and NanoMine data repository in accordance with their licenses and terms of use.

\section{Computational Experiments Details}
\paragraph{Models Details}
All of the open-sourced models used in our experiments (e.g. LLaMA2, LongChat, and Vicuna) have $7$ billion parameters.

\paragraph{Computational Budget}
We perform all of the experiments with one NVIDIA RTX A6000 GPU. Each of the experiments with LLaMA2, LongChat, and Vicuna took $2-3$ hours.

\begin{table}[th]
\centering
\begin{tabular}{ccr}
\toprule
\textbf{$\alpha$} & \textbf{$\#$Predictions} & \textbf{F$_1$} \\
\midrule
\multirow{6}{*}{2} & 9 & 39.3 \\
& 8 & 39.2 \\
& 7 & 41.2 \\
& 6 & 40.8 \\
& 5 & 41.4 \\
& 4 & 39.9 \\
\midrule
\multirow{5}{*}{3} & 9 & 41.8 \\
 & 8 & \textbf{43.4} \\
& 7 & 39.7 \\
& 6 & 39.0 \\
& 5 & 36.0 \\
\bottomrule
\end{tabular}
\caption{F1 scores for alpha levels 2 and 3, with various numbers of predictions.}
\label{tab:f1_scores}
\end{table}

\paragraph{Hyperparameter Settings}
\label{app:sc}
For all experiments, except those involving self-consistency, the temperature parameter is set to zero to ensure consistent evaluation of the models. In the case of the self-consistency experiment, we determine the optimal value for the $\alpha$ threshold by tuning $\alpha$ on the validation set. Table~\ref{tab:f1_scores} shows that the optimal performance is achieved with $\alpha$ at $3$ and by sampling $8$ predictions.

\section{Model Performance on Condensed and Full Papers}
\label{app:all_results}
Table~\ref{tab:results} presents an evaluation of various LLMs across different condensation levels and their performance on full-length papers.

\begin{table*}[ht]
\centering
\small

\begin{tabular}{@{}lrrrrrrrr@{}}
\toprule
Model & \multicolumn{3}{c}{Strict} & \multicolumn{3}{c}{Partial} \\ 
      & Prec. & Rec.& F$_1$ & Prec.& Rec.& F$_1$  \\
\midrule
\multicolumn{7}{c}{Condensed Papers (Top 5)} \\
\midrule
LLaMA2-7b Chat  & 9.4 & 0.4 & 0.7   & 41.5 & 0.9 & 1.8 \\     
LongChat-7b-13k & 3.7 & 1.4 & 2.1   & 43.3 & 15.2 & 22.5  \\   
Vicuna-7b-v1.5  & 5.8 & 2.6 & 3.6   & 49.9 & 19.5 & 28.1  \\   
Vicuna-7b-v1.5-16k  & 17.7 & 5.9 & 8.9   & 60.4 & 19.9 & 29.9  \\ 
GPT-4 Turbo   & 31.9 & 18.6 & 23.5   & 63.1 & 35.6 & 45.5  \\  
\midrule
\multicolumn{7}{c}{Condensed Papers (Top 10)} \\
\midrule
LLaMA2-7b Chat      & 21.7 & 0.6 & 1.2   & 60.0 & 1.5 & 3.0            \\
LongChat-7b-13k     & 2.0 & 0.8 & 1.1   & 45.0 & 17.6 & 25.3        \\
Vicuna-7b-v1.5     & 14.7 & 3.0 & 5.0  & 60.0 & 10.4 & 17.7        \\
Vicuna-7b-v1.5-16k      & 15.0 & 4.9 & 7.4   & 58.3 & 17.8 & 27.3        \\
GPT-4 Turbo      & 33.7 & 23.0 & 27.3  & 61.5 & 42.3 & 50.1  \\
\midrule
\multicolumn{7}{c}{Condensed Papers (Top 30)} \\
\midrule
LongChat-7b-13k & 4.7 & 7.0 & 3.5 & 48.2 & 24.3 & 32.4  \\     
Vicuna-7b-v1.5  & 6.5 & 0.2 & 0.5 & 55.7 & 1.8 & 3.6  \\ 
Vicuna-7b-v1.5-16k  & 17.3 & 5.6 & 8.4 & 62.2 & 18.3 & 28.2  \\ 
GPT-4 Turbo   & 43.6 & \textbf{32.0} & \textbf{36.9}  & 64.5 & \textbf{47.7} & \textbf{54.8}  \\    
\midrule
\multicolumn{7}{c}{Full Papers} \\
\midrule
Vicuna-7b-v1.5-16k   & 18.4 & 1.5 & 2.7   & 65.7 & 4.6 & 8.5                   \\
LongChat-7b-13k   & 5.4 & 4.2 & 4.7   & 36.6 & 29.6 & 32.7  \\
GPT-4 Turbo        & \textbf{44.8} & 30.2 & 36.0   & \textbf{64.9} & 43.8 & 52.3
\\
\bottomrule
\end{tabular}
\caption{Precision, Recall, and F$_1$ of different LLMs on condensed and full papers using strict and partial metrics. The results are segmented based on the degree of paper condensation (Top 5, Top 10, Top 30 segments) and for full paper length
}
\label{tab:results}
\end{table*}

\section{Prompts}
In this section, we present all the prompts used in our experiments.
\subsection{E2E Prompt}
\begin{lstlisting}
Please read the following paragraphs, find all the nano-composite samples, and then fill out the given JSON template for each one of those nanocomposite samples. If there are multiple Filler Composition Mass/Volume for a unique set of Matrix/Filler Chemical Name, please give a list for the Composition. If an attribute is not mentioned in the paragraphs fill that section with "null". Mass and Volume Composition should be followed by a %.

{
    "Matrix Chemical Name": "chemical_name",
    "Matrix Chemical Abbreviation": "abbreviation",
    "Filler Chemical Name": "chemical_name",
    "Filler Chemical Abbreviation": "abbreviation",
    "Filler Composition Mass": "mass_value",
    "Filler Composition Volume": "volume_value"
}

[PAPER SPLIT]
\end{lstlisting}

\subsection{NER prompt}
\begin{lstlisting}
Please identify the matrix name(s), filler name(s), and filler composition fraction(s). Here is an example of what you should return:

{
    "Matrix Chemical Names": ["Poly(vinyl acetate)", "Glycerol"],
    "Matrix Chemical Abbreviation": ["PVAc"],
    "Filler Chemical Names": ["Silicon dioxide"],
    "Filler Chemical Abbreviation": ["SiO2"],
    "Filler Composition Fraction": ["6%", "12%", "20%", "23%", "32%"]
}

[PAPER SPLIT]
\end{lstlisting}

\subsection{RE Prompt}
\begin{lstlisting}
Is the following sample a valid polymer nanocomposite sample mentioned in the article? Yes or No?

Sample: 
[JSON OBJECT]

Article:
[PAPER SPLIT]
\end{lstlisting}

\newpage
\section{Re-Annotation Example Text}
\label{app:re-annot}
Below, we provide an example of the text that is automatically generated which facilitates the re-annotation.

\begin{lstlisting}
File name: L381

True sample 0 is matched with predicted sample 0
But there's a discrepancy between the predicted sample and the true sample Filler Composition Volume.
True sample: {'Matrix Chemical Name': 'Polystyrene', 'Matrix Abbreviation': 'PS', 'Filler Chemical Name': 'Reduced graphene oxide', 'Filler Abbreviation': 'rGO', 'Filler Composition Mass': None, 'Filler Composition Volume': '0.00428'}
Predicted sample: {'Matrix Chemical Name': 'Polystyrene', 'Matrix Chemical Abbreviation': 'PS', 'Filler Chemical Name': 'Reduced Graphene Oxide', 'Filler Chemical Abbreviation': 'rGO', 'Filler Composition Mass': 'null', 'Filler Composition Volume': '2.10%'}

True sample 5 is matched with predicted sample 5
But there's a discrepancy between the predicted sample and the true sample Filler Composition Volume.
True sample: {'Matrix Chemical Name': 'Polystyrene', 'Matrix Abbreviation': 'PS', 'Filler Chemical Name': 'Reduced graphene oxide', 'Filler Abbreviation': 'rGO', 'Filler Composition Mass': None, 'Filler Composition Volume': '0.0127'}
Predicted sample: {'Matrix Chemical Name': 'Polystyrene', 'Matrix Chemical Abbreviation': 'PS', 'Filler Chemical Name': 'Reduced Graphene Oxide', 'Filler Chemical Abbreviation': 'rGO', 'Filler Composition Mass': 'null', 'Filler Composition Volume': '0.053%'}
Standardized predicted sample: {'Matrix Chemical Name': 'Polystyrene', 'Matrix Chemical Abbreviation': 'PS', 'Filler Chemical Name': 'Reduced Graphene Oxide', 'Filler Chemical Abbreviation': 'rGO', 'Filler Composition Mass': 'null', 'Filler Composition Volume': '0.053%'}

True sample 1 is exactly matched with predicted sample 3.

True sample 2 is exactly matched with predicted sample 2. 

True sample 3 is exactly matched with predicted sample 1. 

True sample 4 is exactly matched with predicted sample 4.

\end{lstlisting}

\end{document}